\documentclass[11pt,a4paper]{article}
\usepackage[hyperref]{acl2017}
\usepackage{times}
\usepackage{url}
\usepackage{latexsym}
\usepackage{amsfonts}
\usepackage{graphicx}
\usepackage{booktabs}
\usepackage{rotating}
\usepackage{graphicx,multirow}
\usepackage{makecell}
\aclfinalcopy % Uncomment this line for the final submission
\title{A Simple and Accurate Syntax-Agnostic Neural Model for Dependency-based Semantic Role Labeling}

\author{
Diego Marcheggiani$^1$, ~ Anton Frolov$^2$, ~ Ivan Titov$^{1,3}$ \\
$^1$ILLC, University of Amsterdam \\ 
   $^2$Machine Intelligence Department, Yandex  \\
   $^3$ILCC, School of Informatics, University of Edinburgh \\
  {\tt marcheggiani@uva.nl} \\
  {\tt anton-fr@yandex-team.ru} \\ 
  {\tt ititov@inf.ed.ac.uk}
}

\date{}

\begin{document}
\maketitle
\begin{abstract}
We introduce a simple and accurate neural model for dependency-based semantic role labeling.
Our model predicts predicate-argument dependencies relying on states of a bidirectional LSTM encoder. 
The semantic role labeler achieves competitive performance on 
English, even without any kind of syntactic information and only using local inference. 
However, when automatically predicted part-of-speech tags are provided as input, it substantially outperforms all previous local models and approaches the best reported results on the English CoNLL-2009 dataset. 
We also consider Chinese, Czech and Spanish where our approach also achieves competitive results.
Syntactic parsers are unreliable on out-of-domain data, so standard (i.e., syntactically-informed) SRL models are hindered when tested in this setting. 
Our syntax-agnostic model appears more robust, resulting in the best reported results on standard out-of-domain test sets.
\end{abstract}
%------------------------------------------------------------------------

\section{Introduction}
\begin{figure}
\begin{center}
\includegraphics[width=0.8\columnwidth]{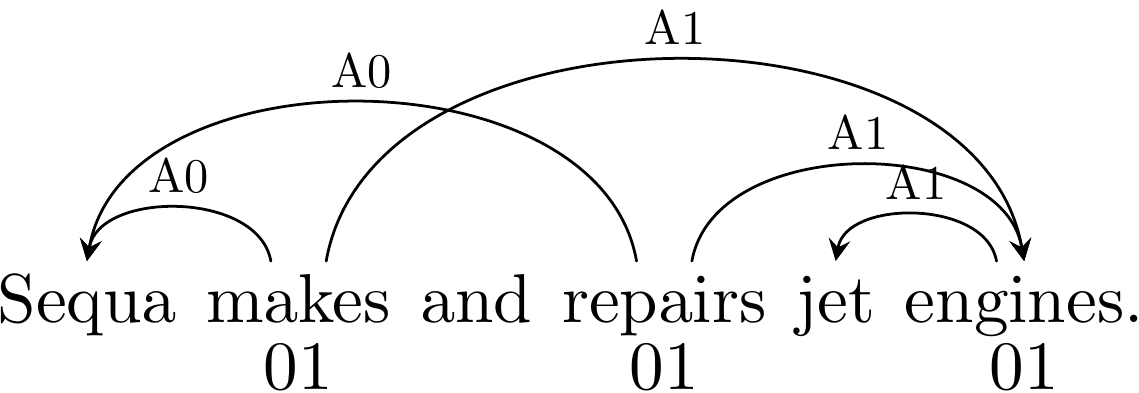}
\caption{\label{fig:example} A semantic dependency graph.
} 
\end{center}
\end{figure}

The task of semantic role labeling (SRL), pioneered by \citet{gildea2002automatic}, 
involves the prediction of predicate argument structure, i.e., both identification of
arguments as well as their assignment to an underlying \textit{semantic role}. 
These representations have been shown to be beneficial in many NLP applications, including question answering~\citep{DBLP:conf/emnlp/ShenL07} and information extraction \citep{DBLP:conf/kcap/ChristensenMSE11}.
Semantic banks (e.g., PropBank~\citep{DBLP:journals/coling/PalmerKG05}) often represent
arguments as syntactic constituents or, more generally, text spans~\citep{DBLP:conf/acl/BakerFL98}. In contrast, CoNLL-2008 and 2009 shared tasks~\citep{DBLP:conf/conll/SurdeanuJMMN08,DBLP:conf/conll/HajicCJKMMMNPSSSXZ09} popularized {\it dependency-based semantic role labeling} where the goal is to identify syntactic heads of arguments rather than entire constituents.
Figure~\ref{fig:example} shows an example of such a dependency-based representation: node labels are senses of predicates (e.g., ``01" indicates that the first sense from the PropBank sense repository is used for predicate {\it makes} in this sentence) and edge labels are semantic roles (e.g., {\tt A0} is a proto-agent, `doer').

Until recently, state-of-the-art SRL systems relied on complex sets of lexico-syntactic features~\citep{DBLP:conf/conll/PradhanHWMJ05} as well as declarative constraints~\citep{DBLP:journals/coling/PunyakanokRY08,DBLP:conf/icml/RothY05}. 
Neural SRL models instead exploited feature induction capabilities of neural networks,
largely eliminating the need for complex hand-crafted features. Initially achieving state-of-the-art results only in the multilingual setting, where careful feature engineering is not practical~\citep{TitovCoNLL09ST,TitovIjcai09},  neural SRL models
 now also outperform their traditional counterparts on standard benchmarks for English~\citep{fitzgerald-EtAl:2015:EMNLP,roth-lapata:2016:P16-1,DBLP:conf/conll/SwayamdiptaBDS16,foland2015}.

Recently, it has been shown that an accurate
span-based SRL model can be constructed without
relying on syntactic features~\citep{zhou-xu:2015:ACL-IJCNLP}.
Nevertheless, 
the situation with dependency-based SRL has not changed: even recent state-of-the-art methods for this task  heavily rely on syntactic features \citep{roth-lapata:2016:P16-1,fitzgerald-EtAl:2015:EMNLP,lei-EtAl:2015:NAACL-HLT,DBLP:conf/emnlp/RothW14,DBLP:conf/conll/SwayamdiptaBDS16}. 
In particular, \citet{roth-lapata:2016:P16-1} argue that syntactic features are necessary for the dependency-based SRL and show that performance of their model degrades dramatically if syntactic paths between arguments and predicates are not provided as an input. In this work, we are the first  to show that it is possible to construct a very accurate dependency-based semantic role labeler which either does not use any kind of syntactic information or uses very little (automatically predicted part-of-speech tags). This suggests that our LSTM model can largely implicitly capture syntactic information, and this information can, to a large extent, substitute treebank syntax.

Similarly to the span-based model of \citet{zhou-xu:2015:ACL-IJCNLP} we use bidirectional LSTMs to encode sentences and rely on their states when predicting arguments of each predicate.\footnote{In the CoNLL-2009 benchmark, predicates do not need to be identified: their positions are provided as input at test time. Consequently, as standard for dependency SRL, we ignore this subtask in further discussion.}  We predict semantic dependency edges between predicates and arguments relying on LSTM states corresponding to the predicate and the argument positions (i.e. both edge endpoints). 
As semantic roles are often specific to predicates or even  predicate senses (e.g., in PropBank~\cite{DBLP:journals/coling/PalmerKG05}), instead of predicting the role label (e.g., A0 for {\it Sequa} in our example), we predict predicate-specific roles (e.g., {\it make}-A0) using a compositional model. Both these aspects (predicting edges and compositional embeddings of roles) contrast our approach with that of \newcite{zhou-xu:2015:ACL-IJCNLP} who essentially treat the SRL task as a generic sequence labeling task. We empirically show that using these two ideas is crucial for achieving competitive performance on dependency SRL (+1.0\% semantic {F\textsubscript{1}} in our ablation studies on English). Also, unlike the span-based version, we observe that using automatically predicted POS tags is also important (+0.7\%~{F\textsubscript{1}}).

The resulting SRL model is very simple. 
Not only we do not rely on syntax, our model is also local, i.e., we do not globally score or constrain sets of arguments. 
On the standard English in-domain CoNLL-2009 benchmark we achieve $87.7$~{F\textsubscript{1}} which compares favorable to the best local model ($86.7$\% {F\textsubscript{1}} for PathLSTM~\citep{roth-lapata:2016:P16-1}) and approaches the best results overall ($87.9$\% for an ensemble of 3 PathLSTM models with a reranker on top). 
When we experiment with Chinese, Czech and Spanish portions of the CoNLL-2009 dataset, we also achieve competitive results, even without any extra hyper-parameter tuning.

Moreover, as syntactic parsers are not reliable when used out-of-domain,
standard (i.e., syntactically-informed) dependency SRL models are crippled when applied to such data. 
In contrast, our syntax-agnostic model appears to be considerably more robust: we achieve the best result so far on the English and Czech out-of-domain test set ($77.7$\% and $87.2$\% {F\textsubscript{1}}, respectively).
For English, this constitutes a $2.4$\% absolute improvement over the comparable previous model ($75.3$\% for the local PathLSTM) and substantially outperforms any previous method ($76.5$\% for the ensemble of 3 PathLSTMs). We believe that out-of-domain performance may in fact be more important than in-domain one: in practice  linguistic tools are rarely, if ever, used in-domain.

The key contributions can be summarized as follows:
\begin{itemize}
\item we propose the first effective syntax-agnostic model for dependency-based SRL;
\item it achieves the best results among local models on the English, Chinese and Czech in-domain test sets;
\item it substantially outperforms all previous methods on the out-of-domain test set on both English and Czech.
\end{itemize}
Despite the effectiveness of our syntax-agnostic version, we believe that both integration of treebank syntax and global inference are promising directions and

leave them for future work. In fact, the proposed SRL model, given its simplicity and efficiency, may be used as a natural building block for future global and syntactically-informed SRL models.\footnote{The code is available at \url{https://github.com/diegma/neural-dep-srl}.}

\section{Our Model}
The focus of this paper is on argument identification and labeling, as these are the steps which have been previously believed to require syntactic information. 
For the predicate disambiguation subtask we use   
models from previous work.

In order to identify and classify arguments, we propose a model composed of three components:
\begin{itemize}
\item  a word representation component that from a word $w_i$ in a sentence $\mathbf{w}$ build a word representation $x_i$;
\item  a Bidirectional LSTM (BiLSTM) encoder which takes as input the word  representation $x_i$ and provide a dynamic representation of the word and its context in a sentence;
\item a classifier which takes as an input the BiLSTM representation of the candidate argument and the BiLSTM representation of the predicate to predict the role associated to the candidate argument.
\end{itemize}

\begin{figure}
\begin{center}
\includegraphics[width=\columnwidth]{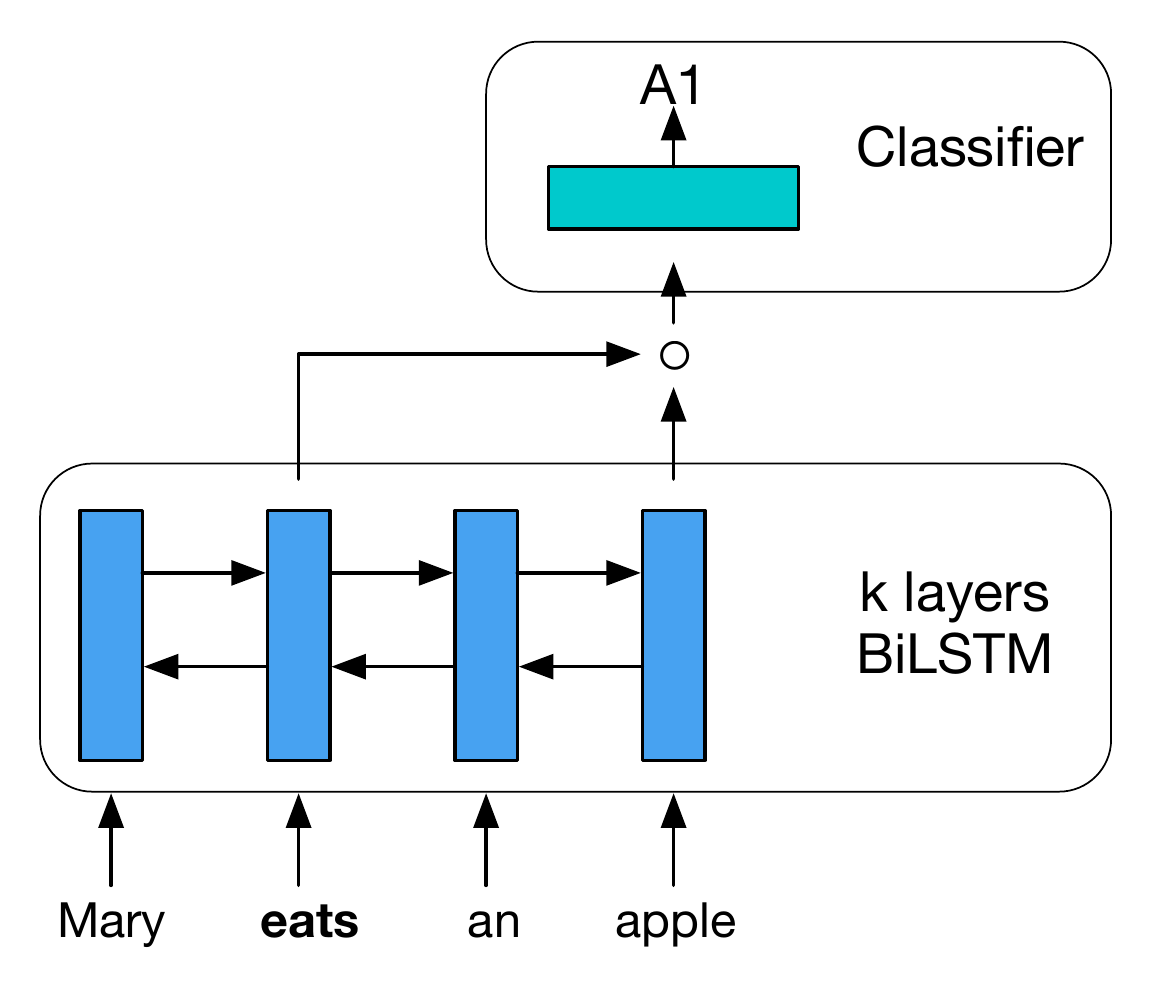}
\caption{\label{fig:model} Predicting an argument and its label with an LSTM encoder. 
} 
\end{center}
\end{figure}

\subsection{Word representation} \label{sec:word-repr}
We represent each word $w$ as the concatenation of four vectors: a randomly initialized word embedding $x^{re} \in \mathbb{R}^{d_w}$, a pre-trained word embedding $x^{pe} \in \mathbb{R}^{d_w}$, a randomly initialized  part-of-speech tag embedding $x^{pos} \in \mathbb{R}^{d_p}$ and a randomly initialized lemma embedding $x^{le} \in \mathbb{R}^{d_l}$ that is only active if the word is one of the predicates.
The randomly initialized embeddings $x^{re}$, $x^{pos}$, and $x^{le}$ are fine-tuned during training, while the pre-trained ones are kept fixed, as in \citet{dyer-EtAl:2015:ACL-IJCNLP}.
The final word representation is given by $x = x^{re} \circ x^{pe} \circ x^{pos} \circ x^{le}$, where $\circ$ represents the concatenation operator.

\subsection{Bidirectional LSTM encoder}
One of the most effective ways to model sequences are recurrent neural networks (RNN) \citep{DBLP:journals/cogsci/Elman90}, more precisely their gated versions, for example, Long Short-Term Memory (LSTM) networks \citep{DBLP:journals/neco/HochreiterS97}.

Formally, we can define an LSTM as a function $LSTM_{\theta}(x_{1:i})$ that takes as input the sequence $x_{1:i}$ and returns a hidden state $h_i \in \mathbb{R}^{d_h}$. 
This state can be regarded as a representation of the sentence from the start to the position $i$, or, in other words, it encodes the word at position $i$ along with its left context. 
Bidirectional LSTMs make use of two LSTMs: one for the forward pass, and another for the backward pass, $LSTM_{F}$ and $LSTM_{B}$, respectively. In this way the concatenation of forward and backward LSTM states encodes both left and right contexts of a word, $
BiLSTM(x_{1:n},i) = LSTM_{F}(x_{1:i}) \circ LSTM_{B}(x_{n:i})$.
In this work we stack $k$ layers of bidirectional LSTMs, each layer takes the lower layer as its input.

\subsection{Predicate-specific encoding}\label{sec:pred-flag}

As we will show in the ablation studies in Section \ref{sec:experiments}, encoding a sentence with a bidirectional LSTM in one shot and using it to predict the entire semantic dependency graph does not result in competitive SRL performance. Instead, similarly to \citet{zhou-xu:2015:ACL-IJCNLP}, we produce predicate-specific encodings of a sentence and use them to predict arguments of the corresponding predicate. 
This contrasts with most other applications of LSTM encoders (for example, in syntactic parsing~\cite{TACL885,cross-huang:2016:P16-2} or machine translation~\cite{DBLP:conf/nips/SutskeverVL14}), where sentences are typically encoded once and then used to predict the entire structured output (e.g., a syntactic tree or a target sentence). 

Specifically, when identifying arguments of a given predicate, we add a predicate-specific feature to the  representation of each word in the sentence by concatenating a binary flag to the word representation of Section \ref{sec:word-repr}.
The flag is set to 1 for the word corresponding to the currently considered predicate, it is set to 0 otherwise.
In this way, sentences with more than one predicate will be re-encoded by bidirectional LSTMs multiple times.

\subsection{Role classifier}

Our goal is to predict and label arguments for a given predicate. This can be accomplished by labeling each word in a sentence with a role, including the special `NULL' role to indicate that it is not an argument of the predicate. We start with explaining the basic  role classifier and then discuss two extensions, which we will later show to be crucial for achieving competitive performance. 

\subsubsection{Basic role classifier}
\label{sec:basic}

The basic role classifier takes the hidden state of the top-layer bidirectional LSTM  corresponding to the  considered word at position $i$ and uses it to estimate the probability of the role $r$. Though we experimented with multilayer perceptrons, we obtained the best results with a simple log-linear model:
\begin{equation} \
\label{eq:basic-softmax}
p(r|v_i, p) \propto \exp(W_{r} v_i ),
\end{equation}
where $v_i$ is the hidden state calculated by  $
BiLSTM(x_{1:n},i)$, $p$ refers to the predicate and the symbol $\propto$ signifies proportionality.
This is essentially equivalent to the approach used in \citet{zhou-xu:2015:ACL-IJCNLP} for span-based SRL.\footnote{Since they considered span-based SRL, they used BIO encoding~\cite{ramshaw1995text} and ensured the consistency of B, I and O labels with a 1-order Markov CRF.  For dependency SRL both BIO encoding and the 1-order Markov CRF would be useless.}

\subsubsection{Incorporating predicate state}
\label{sec:pred-state}

Since the context of a predicate in the sentence is highly informative for deciding if a word is its argument and for choosing its semantic role, we provide the predicate's hidden state ($v_p$) as another input to the classifier (as in Figure~\ref{fig:model}):
\begin{equation} \label{eq:softmax}
p(r|v_i, v_p) \propto \exp(W_{r} ( v_i \circ v_p)),
\end{equation}
where, as before, $\circ$ denotes concatenation. 
Note that we are effectively predicting an edge between words $i$ and $p$
in the sentence, so it is quite natural to exploit hidden states corresponding to both endpoints.\footnote{We abuse the notation and refer as $p$ both to the predicate word and to its position in the sentence.} 

Since we use predicate information within the classifier, it may seem that predicate-specific sentence encoding (Section~\ref{sec:pred-flag}) is not needed anymore. Moreover, predicting dependency edges relying on LSTM states of endpoints was shown effective in the context of syntactic dependency parsing without any form of re-encoding~\cite{TACL885}. Nevertheless, in our ablation studies we observed that foregoing predicate-specific encoding results in  large performance degradation (-6.2\% F$_1$ on English). Though  this dramatic drop in performance seems indeed surprising, the nature of the semantic dependencies, especially for nominal predicates, is different from general syntactic dependencies, with many arguments being far away from the predicates. Relations of these arguments to the predicate may be hard to encode with this simpler mechanism.

The two ways of encoding predicate information, using predicate-specific encoding and incorporating the predicate state in the classifier, turn out to be complementary. 

\subsubsection{Compositional modeling of roles}
\label{sec:comp}

Instead of using a matrix $W_{r}$ 
we found it beneficial to jointly embed the role $r$ and predicate lemma $l$ using a non-linear transformation:
\begin{equation}
\label{eq:final-class}
p(r|v_i, v_p, l) \propto \exp(W_{l, r} ( v_i \circ v_p)),
\end{equation}
\begin{equation}
\label{eq:frame_role}
 W_{l,r} =  ReLU(U ({u}_l\circ {v}_r)),% [r x  4h] = [r x cc] [cc x 4h]  
 \end{equation}
where $ReLU$ is the rectilinear activation function, $U$ is a parameter matrix, whereas ${u}_l\in \mathbb{R}^{d'_l}$ and ${v}_r \in \mathbb{R}^{d_r}$ are randomly initialized embeddings of predicate lemmas and roles. 
In this way each role prediction is predicate-specific, and at the same time we expect to learn a good representation for roles associated to infrequent predicates.
This form of compositional embedding is similar to the one used in \citet{fitzgerald-EtAl:2015:EMNLP}.

\section{Experiments}\label{sec:experiments}

We applied our model to the English, Chinese, Czech and Spanish CoNLL-2009 datasets with the standard split into training, test and development sets. 
For English, we used external embeddings of \citet{dyer-EtAl:2015:ACL-IJCNLP} learned using the structured skip n-gram approach of \citet{ling-EtAl:2015:NAACL-HLT}, for Chinese, we used external embeddings produced with the neural language model of \newcite{DBLP:journals/jmlr/BengioDVJ03}.
For Czech and Spanish, we used embeddings created with the model proposed by \newcite{bojanowski2016enriching}.

Similarly to \citet{TACL885} we used word dropout \citep{iyyer-EtAl:2015:ACL-IJCNLP}; we replaced a word with the \textit{UNK} token with probability $\frac{\alpha}{fr(w) + \alpha}$, where $\alpha$ is an hyper-parameter and $fr(w)$ is the frequency of the word $w$.
The predicted POS tags were provided by the CoNLL-2009 shared-task organizers.
We used  the same predicate disambiguator as in \newcite{roth-lapata:2016:P16-1} for English, the one used in \newcite{DBLP:conf/conll/ZhaoCKUT09} for Czech and Spanish, and the one used in \newcite{DBLP:conf/conll/BjorkelundHN09} for Chinese.
The training objective was the categorical cross-entropy, and we optimized it 
 with Adam \citep{kingma2014adam}. 
The hyperparameter tuning and all model selection was performed on the English development set; the chosen values are shown in Table \ref{tab:hyperparameters}.

\begin{table}[h]
\centering
\begin{tabular}{@{\extracolsep{\fill}}l@{\hspace{6pt}}l@{\hspace{6pt}}c}
  \toprule
  &Semantic role labeler& \\
  \midrule
   & $d_w$ (English word embeddings) & 100   \\
   & $d_w$ (Chinese word embeddings) & 128   \\
   & $d_w$ (Czech word embeddings) & 300   \\
   & $d_w$ (Spanish word embeddings) & 300   \\
   & $d_{pos}$ (POS embeddings) &  16  \\
   & $d_l$ (lemma embeddings) & 100 \\
   & $d_h$ (LSTM hidden states) &   512  \\
   & $d_{r}$ (role representation) & 128  \\
   & $d'_{l}$ (output lemma representation) & 128  \\
   & $k$ (BiLSTM depth) & 4  \\
  & $\alpha$ (word dropout) & .25  \\
   & learning rate & .01  \\
\bottomrule
\end{tabular}
\caption{\label{tab:hyperparameters} Hyperparameter values. 
}
\end{table}

\subsection{Results}
\begin{table}[t]
\centering
\scalebox{0.9}{
\begin{tabular}{@{\extracolsep{\fill}}l@{\hspace{6pt}}l@{\hspace{6pt}}c@{\hspace{6pt}}c@{\hspace{6pt}}c}
  \toprule
   & System & P & R & {F\textsubscript{1}}   \\
   \midrule
   &\newcite{lei-EtAl:2015:NAACL-HLT} \small{(local)} & - & - & 86.6 \\
   &\newcite{fitzgerald-EtAl:2015:EMNLP} \small{(local)} & - & - & 86.7 \\
   & \newcite{roth-lapata:2016:P16-1} \small{(local)} & 88.1 & 85.3 & 86.7 \\
& {\bf Ours} \small{\bf (local)} & {\bf 88.7} & {\bf 86.8} & {\bf 87.7} \\
   \midrule
   &\newcite{bjorkelund2010high} \small{(global)} & 88.6 & 85.2 & 86.9 \\
      &\newcite{fitzgerald-EtAl:2015:EMNLP} \small{(global)} & - & - & 87.3 \\
      &\newcite{foland2015} \small{(global)} & - & - & 86.0 \\
  &\newcite{DBLP:conf/conll/SwayamdiptaBDS16} \small{(global)} & - & - & 85.0 \\
   &\newcite{roth-lapata:2016:P16-1} \small{(global)} & 90.0 & 85.5 & 87.7 \\
   \midrule
   &\newcite{fitzgerald-EtAl:2015:EMNLP} \small{(ensemble)} & - & - & 87.7 \\
   &\newcite{roth-lapata:2016:P16-1} \small{(ensemble)} & 90.3 & 85.7 & 87.9 \\
\bottomrule
\end{tabular}
}
\caption{\label{tab:wsj-results} Results on the English in-domain test set. 
}
\end{table}

\begin{table}[t]
\centering
\scalebox{0.9}{
\begin{tabular}{@{\extracolsep{\fill}}l@{\hspace{6pt}}l@{\hspace{6pt}}c@{\hspace{6pt}}c@{\hspace{6pt}}c}
  \toprule
   & System & P & R & {F\textsubscript{1}}   \\
   \midrule
   &\newcite{lei-EtAl:2015:NAACL-HLT} \small{(local)} & - & - & 75.6 \\
   &\newcite{fitzgerald-EtAl:2015:EMNLP} \small{(local)} & - & - & 75.2 \\
   & \newcite{roth-lapata:2016:P16-1} \small{(local)} & 76.9 & 73.8 & 75.3 \\
& {\bf Ours} \small{\bf (local)} & {\bf 79.4} & {\bf 76.2} & {\bf 77.7} \\
   \midrule
   &\newcite{bjorkelund2010high} \small{(global)} & 77.9 & 73.6 & 75.7 \\
   &\newcite{fitzgerald-EtAl:2015:EMNLP} \small{(global)} & - & - & 75.2 \\
   &\newcite{foland2015} \small{(global)} & - & - & 75.9 \\
   &\newcite{roth-lapata:2016:P16-1} \small{(global)} & 78.6 & 73.8 & 76.1 \\
   \midrule
   &\newcite{fitzgerald-EtAl:2015:EMNLP} \small{(ensemble)} & - & - & 75.5 \\
   &\newcite{roth-lapata:2016:P16-1} \small{(ensemble)} & 79.7 & 73.6 & 76.5 \\
\bottomrule
\end{tabular}
}
\caption{\label{tab:ood-results} Results on the English out-of-domain test set. 
}
\end{table}

We compared our full model (with POS tags and the classifier defined in Section~\ref{sec:comp}) against state-of-the-art models for dependency-based SRL on English, Chinese, Czech and Spanish.
For English,  our model significantly outperformed all the local counter-parts (i.e., models which do not perform global inference) on the in-domain tests (see Table~\ref{tab:wsj-results}) with 87.6\% {F\textsubscript{1}} for our model vs. 86.7\% for PathLSTM~\cite{roth-lapata:2016:P16-1}.
When compared with global models, our model performed on-par with the state-of-the-art global version of PathLSTM.

Though we had not done any parameter selection for other languages (i.e., used the same parameters as for English), our model performed competitively across all languages we considered. 

For Chinese (Table~\ref{tab:chinese-results}), the proposed model outperformed the best previous model (PathLSTM) with an improvement of 1.8\% {F\textsubscript{1}}.  

\begin{table}[t]
\centering
\begin{tabular}{@{\extracolsep{\fill}}l@{\hspace{6pt}}l@{\hspace{6pt}}c@{\hspace{6pt}}c@{\hspace{6pt}}c}
  \toprule
   & System & P & R & {F\textsubscript{1}}   \\
   \midrule
   &\newcite{DBLP:conf/conll/BjorkelundHN09}  & 82.4 & 75.1 & 78.6 \\
   &\newcite{DBLP:conf/conll/ZhaoCKUT09}  & 80.4 & 75.2 & 77.7 \\
   &\newcite{roth-lapata:2016:P16-1}  & 83.2 & 75.9 & 79.4 \\
   & {\bf Ours}  & {\bf 83.4} & {\bf 79.1} & {\bf 81.2} \\
\bottomrule
\end{tabular}
\caption{\label{tab:chinese-results} Results on the Chinese test set. 
}
\end{table}

For Czech (Table~\ref{tab:czech-results}), our model, even though unlike previous work it does not use any kind of  morphological features explicitly,\footnote{However, character level information is encoded in the external embeddings, see \cite{bojanowski2016enriching}.} was able to outperform the system that achieved the best score in the CoNLL-2009 shared task. The improvement is 0.8\%~{F\textsubscript{1}}.

\begin{table}[t]
\centering
\begin{tabular}{@{\extracolsep{\fill}}l@{\hspace{6pt}}l@{\hspace{6pt}}c@{\hspace{6pt}}c@{\hspace{6pt}}c}
  \toprule
   & In-domain & P & R & {F\textsubscript{1}}   \\
   \midrule
   &\newcite{DBLP:conf/conll/BjorkelundHN09}  & 88.1 & 82.9 & 85.4 \\
   &\newcite{DBLP:conf/conll/ZhaoCKUT09}  & 88.2 & 82.4 & 85.2 \\
   & {\bf Ours}  & {\bf 86.6} & {\bf 85.4} & {\bf 86.0} \\
   \midrule
   & Out-of-domain & P & R & {F\textsubscript{1}}   \\
   \midrule
   &\newcite{DBLP:conf/conll/BjorkelundHN09}  & 86.1 & 81.9 & 83.9 \\
   &\newcite{DBLP:conf/conll/ZhaoCKUT09}  & 88.6 & 82.5 & 85.4 \\
   & {\bf Ours}  & {\bf 88.0} & {\bf 86.5} & {\bf 87.2} \\
\bottomrule
\end{tabular}
\caption{\label{tab:czech-results} Results on the Czech test sets. 
}
\end{table}

Finally, for Spanish (Table~\ref{tab:spanish-results}), our system, though again achieved competitive results,    did not outperform the best CoNLL-2009 model and yielded  results very similar to those of PathLSTM.
One possible reason for this slightly weaker performance is the relatively small size of the Spanish training set (less then half of the English one). This suggests that our model, tuned on English, is likely  over-parametrized or under-regularized for Spanish. 

\begin{table}[t]
\centering
\begin{tabular}{@{\extracolsep{\fill}}l@{\hspace{6pt}}l@{\hspace{6pt}}c@{\hspace{6pt}}c@{\hspace{6pt}}c}
  \toprule
   & System & P & R & {F\textsubscript{1}}   \\
   \midrule
   &\newcite{DBLP:conf/conll/BjorkelundHN09}  & 78.9 & 74.3 & 76.5 \\
   &\newcite{DBLP:conf/conll/ZhaoCKUT09}  & 83.1 & 78.0 & 80.5 \\
   &\newcite{roth-lapata:2016:P16-1}  & 83.2 & 77.4 & 80.2 \\
   & {\bf Ours}  & {\bf 81.4} & {\bf 79.3} & {\bf 80.3} \\
\bottomrule
\end{tabular}
\caption{\label{tab:spanish-results} Results on the Spanish test set. 
}
\end{table}

The results are especially strong on out-of-domain data.
As shown in Table~\ref{tab:ood-results}, our approach outperformed even ensemble models on the out-of-domain English data (77.7\% vs. 76.5\% for the ensemble of PathLSTMs). Similarly, it performed very well on the out-of-domain Czech dataset scoring 87.2\% {F\textsubscript{1}}, with a 1.8\% {F\textsubscript{1}} improvement over the best CoNLL-2009 participant (see Table~\ref{tab:czech-results}, bottom). 
The favorable results on out-of-domain test sets are not surprising, as
syntactic parsers, even the most accurate ones, usually struggle on domains different from the ones they have been trained on. 
This means that the syntactic trees they produce are unreliable and compromise the accuracy of SRL systems which rely on them. 
The error propagation can in principle be mitigated by exploiting a distribution over parse trees (e.g., encoded in a parse forest) rather than using a single ('Viterbi') parse. 
However, this is rarely feasible in practice.
Since our model does not use predicted parse trees and instead relies on  the ability of LSTMs to capture long distance dependencies and syntactic phenomena \cite{DBLP:journals/tacl/LinzenDG16}, it is less brittle in this setting.

\subsection{Ablation studies and analysis}
\begin{table}[t]
\centering
\begin{tabular}{@{\extracolsep{\fill}}l@{\hspace{6pt}}l@{\hspace{6pt}}c@{\hspace{6pt}}c@{\hspace{6pt}}c}
  \toprule
   & System & P & R & {F\textsubscript{1}}   \\
   \midrule
   & Ours \small{(local)} & 87.7 & 85.5 & 86.6 \\
   \midrule
   &  w/o POS tags  & 87.3 & 84.5 & 85.9 \\
   &  w/o predicate-specific encoding  & 80.9 & 79.8 & 80.4 \\
   &  with basic classifier  & 86.7 & 84.5 & 85.6 \\
\bottomrule
\end{tabular}
\caption{\label{tab:ablation} Ablation study on the English development set.}
\end{table}
In order to show the contribution of the modeling choices we made, we performed an ablation study on the English development set (Table~\ref{tab:ablation}). 
In these experiments we made individual changes to the model (one by one)  and measured their influence on the model performance.

First, we observed that  POS tag information is highly beneficial for obtaining competitive performance. 

Not using predicate-specific encoding (Section~\ref{sec:pred-flag}), or, in other words, doing one-pass encoding with no predicate flags, hurts the performance even more badly (6\% drop in {F\textsubscript{1}} on the development set). This is somewhat surprising given that one-pass LSTM encoders  performed competitively for syntactic dependencies~\cite{TACL885,cross-huang:2016:P16-2} and suggests that major differences between the two problems require the use of different modeling approaches.

We also observed a 1.0\% drop in {F\textsubscript{1}} when we follow \citet{zhou-xu:2015:ACL-IJCNLP} and use the basic role classifier (Section~\ref{sec:basic}). These results show that both predicate-specific encoding (Section~\ref{sec:pred-flag}) and exploiting predicate information in the classifier (Sections~\ref{sec:pred-state}-\ref{sec:comp}) are complementary.

\begin{figure}
\begin{center}
\includegraphics[width=\columnwidth]{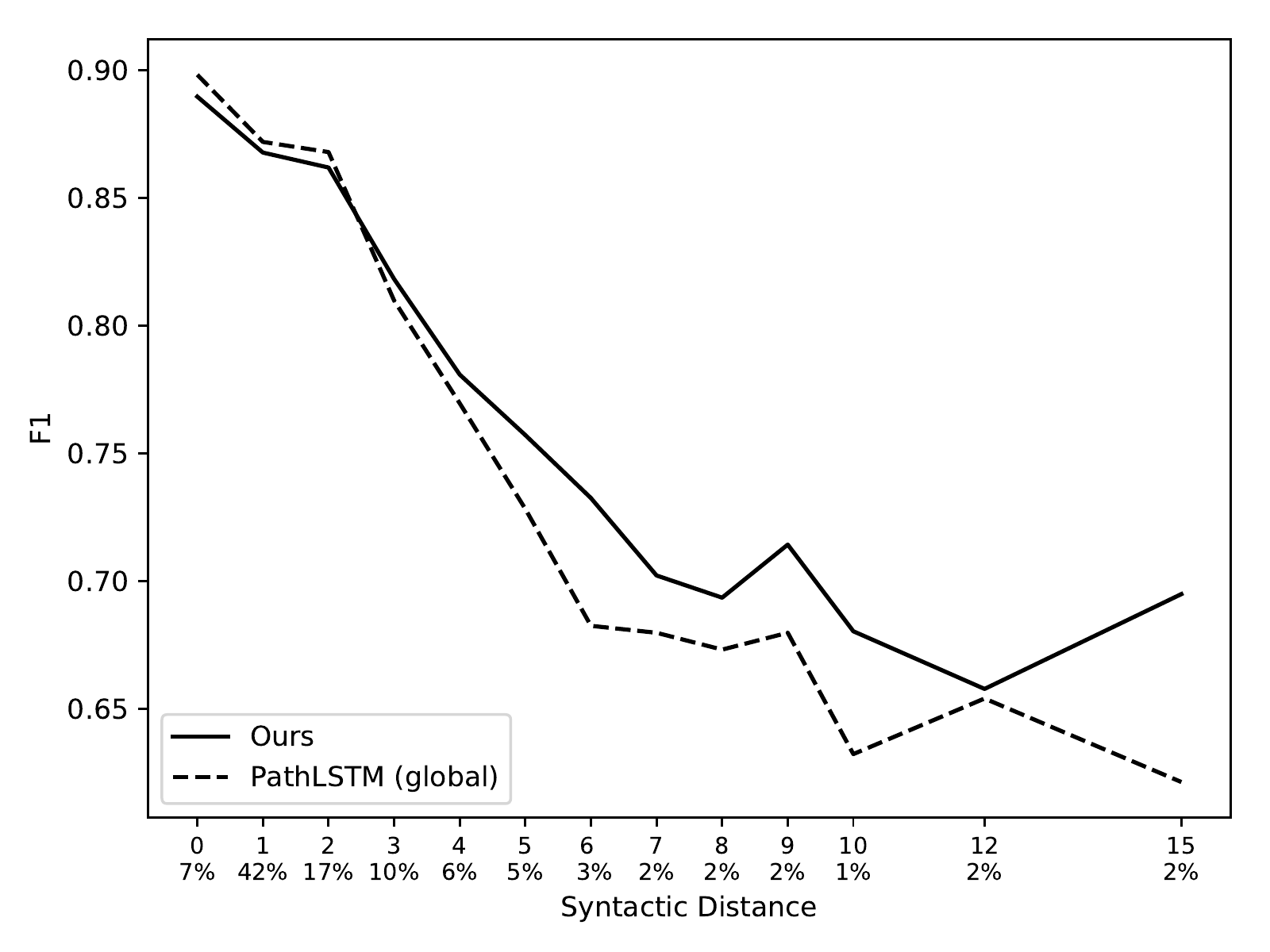}
\caption{\label{fig:token_dist} $F_1$ as function of word distance. Percentages indicate the amount of arguments at a specific distance from a predicate.
} 
\end{center}
\end{figure}

We also studied how performance varies depending on the distance between a predicate and an argument (Figure~\ref{fig:token_dist}). We compared our approach to the global PathLSTM model: PathLSTM is a natural reference point as it is the most accurate previous model, exploits similar modeling and representation techniques (e.g., word embeddings, LSTMs) but, unlike our approach, relies on predicted syntax.
Contrary to our expectations, syntactically-driven and global PathLSTM was weaker for longer distances.  
We may speculate that syntactic paths for arguments further away from the predicate become unreliable. 
Though LSTMs are likely to be affected by a similar trend, their states may be able to capture the uncertainty about the structure and thus let the role classifier account for this uncertainty without the need to explicitly sum over potential syntactic analysis. 
In contrast, PathLSTM will have access only to the single (top scoring) parse tree and, thus, may be more brittle.

\begin{table}[t]
\centering
\begin{tabular}{@{\extracolsep{\fill}}l@{\hspace{6pt}}l@{\hspace{6pt}}l@{\hspace{6pt}}c@{\hspace{6pt}}c@{\hspace{6pt}}c@{\hspace{6pt}}c@{\hspace{6pt}}c}
  \toprule
  & & &  Ours & PathLSTM  & Freq. (\%)\\
   \midrule
   &\multirow{5}{*}{\rotatebox[origin=c]{90}{Verbal}}& A0  &  90.5&  90.4 & 15\%  \\
   && A1  &  92.0&  91.8 & 21\% \\
   && A2  &  80.3&  80.2 & 5\% \\
   && AM-*  & 77.9 & 77.0 & 16\% \\
   && {\it All}  & {\it 86.4} & {\it 86.1} & {\it 61\%} \\
   \midrule
   & &  & Ours & PathLSTM & Freq. (\%)  \\
   \midrule
   &\multirow{5}{*}{\rotatebox[origin=c]{90}{Nominal}}& A0  & 81.8 &  81.5 & 10\% \\
   && A1  & 85.1 &  85.5 & 16\% \\
   & & A2  & 78.5 &  79.8 & 7\% \\
   && AM-*  & 72.5 & 73.2 & 5\% \\
   && {\it All}  & {\it 81.1} & {\it 81.8} & {\it 39\%} \\
\bottomrule
\end{tabular}
\caption{\label{tab:english-nominal-verbal} {F\textsubscript{1}} results on the English test set broken down into verbal and nominal predicates.}
\end{table}

\begin{figure}
\begin{center}
\includegraphics[width=\columnwidth]{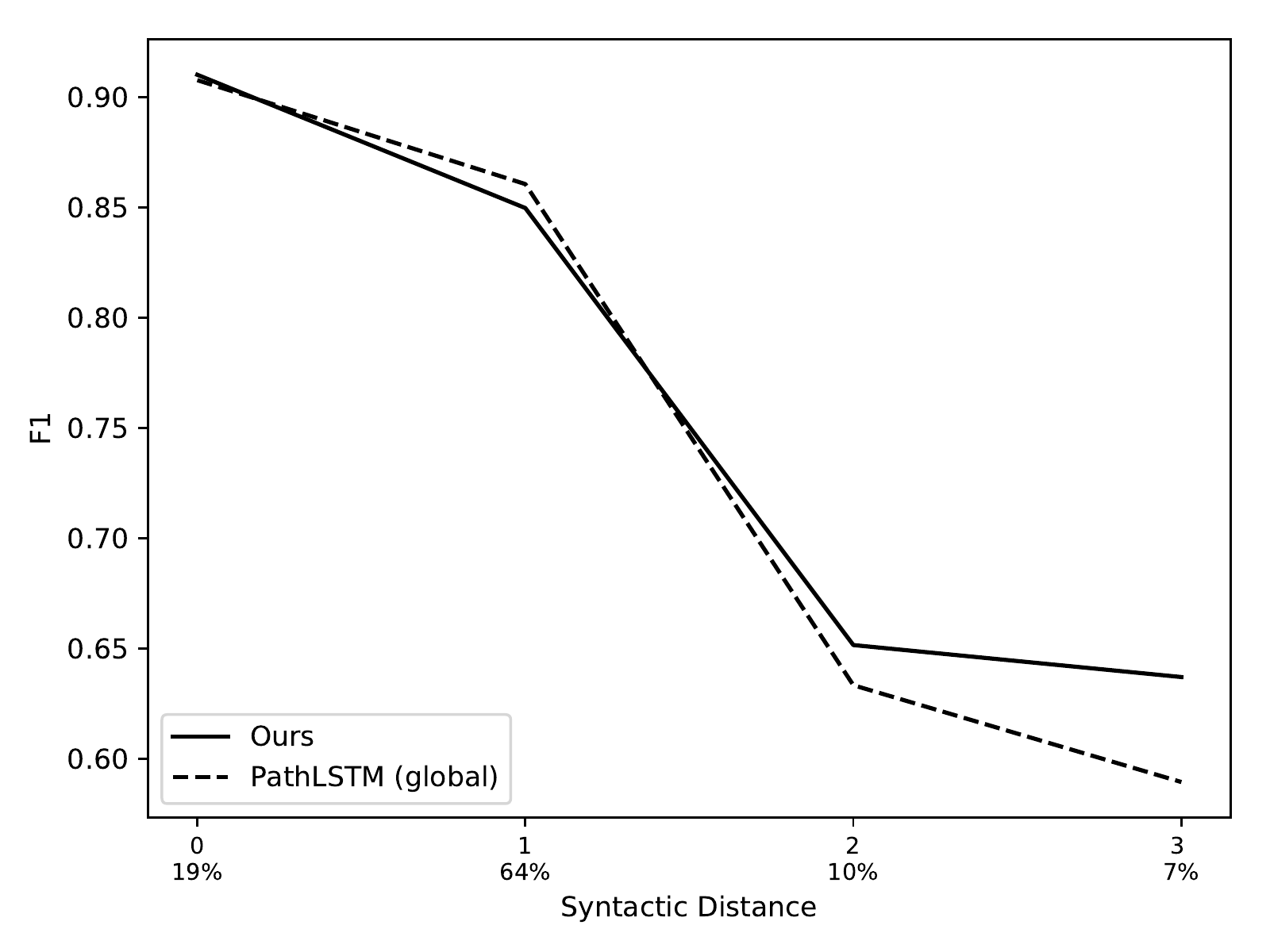}
\caption{\label{fig:dep_dist_nominal} $F_1$ as function of syntactic distance for nominal predicates.  Percentages indicate the amount of arguments at a specific distance from a nominal predicate.
} 
\end{center}
\end{figure}

In Table \ref{tab:english-nominal-verbal}, we break down  {F\textsubscript{1}} results on the English test set into  verbal and nominal predicates, and again compare our results with PathLSTM.
First, as expected, we observe that both models are less accurate in predicting semantic roles of nominal predicates. 
For verbal predicates, our model slightly outperformed PathLSTM in core roles (A0-2) and performed much better (0.9\% {F\textsubscript{1}}) in predicting modifiers (AM-*). This is very surprising as some information about modifiers is actually explicitly encoded in syntactic dependencies exploited by PathLSTM (e.g., the syntactic  dependency TMP is predictive of the modifier role AM-TMP).
Note though that the syntactic parser was trained on the same sentences (both data originates from WSJ sections 02-22 of Penn Treebank), and this can explain why these syntactic dependencies (e.g., TMP) may convey little beneficial information to the semantic role labeler.
For nominal predicates, PathLSTM was more accurate than our model for all roles excluding A0.
To get a better idea for what is happening, we plotted the {F\textsubscript{1}} scores as a function of the length of the shortest path between nominal predicates and their arguments.
On one hand, Figure \ref{fig:dep_dist_nominal} shows that PathLSTM is more accurate on roles one syntactic arc away from the nominal predicate.
Note that these are the majority (78\%) of arguments. 
On the other hand, our model appears to be more accurate for arguments syntactically far from nominal predicates. This again suggests that PathLSTM struggles with harder cases.

\begin{table}[t]
\centering
\begin{tabular}{@{\extracolsep{\fill}}l@{\hspace{6pt}}l@{\hspace{6pt}}l@{\hspace{6pt}}c@{\hspace{6pt}}c@{\hspace{6pt}}c}
  \toprule
   && System & P & R & {F\textsubscript{1}}   \\
   \midrule
   & \multirow{2}{*}{\rotatebox[origin=c]{90}{Verbal}}& PathLSTM & 93.4 & 87.8 & 90.5 \\
   & & {\bf Ours}  & {\bf 92.7} & {\bf 89.8} & {\bf 91.2} \\
   \midrule
    && System & P & R & {F\textsubscript{1}}   \\
   \midrule
   & \multirow{2}{*}{\rotatebox[origin=c]{90}{Nom.}}& PathLSTM  & 92.0 & 83.9 & 87.8 \\
   & & {\bf Ours}  & {\bf 89.4} & {\bf 86.6} & {\bf 88.0} \\
\bottomrule
\end{tabular}
\caption{\label{tab:arg-recognition} Argument recognition results broken down into verbal and nominal predicates. 
}
\end{table}

\begin{table*}[t]
\centering
\begin{tabular}{@{\extracolsep{\fill}}l@{\hspace{6pt}}l@{\hspace{6pt}}l@{\hspace{6pt}}l@{\hspace{6pt}}c@{\hspace{6pt}}c}
  \toprule
   & System & Example   \\
   \midrule
   &Manual& \makecell[l]{Most of the stock \textsubscript{[A2]}selling\textsubscript{[\textbackslash A2]} {\bf pressure} came \textsubscript{[A0]}from\textsubscript{[\textbackslash A0]} Wall Street professionals.}\\
   \midrule
   &PathLSTM& \makecell[l]{Most of the stock \textsubscript{[A2]}selling\textsubscript{[\textbackslash A2]} {\bf pressure} came from  Wall Street professionals.} \\
   \midrule
   &Ours&  \makecell[l]{Most of the stock \textsubscript{[A0]}selling\textsubscript{[\textbackslash A0]} {\bf pressure} came \textsubscript{[A0]}from\textsubscript{[\textbackslash A0]} Wall Street professionals.} \\
\bottomrule
\end{tabular}
\caption{\label{tab:nominal-examples} Example of errors for the nominal predicate {\it pressure}: A0 is {\it a presser} (proto-agent) and A2 is {\it a goal
}. 
}
\end{table*}

Unlike verbal predicates, syntactic structure is less predictive of 
semantic roles for nominals (e.g., many arguments are noun modifiers). Consequently, we hypothesized that our model should be weaker than PathLSTM in recognizing arguments but should be on par with PathLSTM in assigning their roles. To test this, we looked into argument identification performance (i.e., ignored labels).
Table \ref{tab:arg-recognition} shows the accuracy of both models in recognizing arguments of nominal and verbal predicates.
Our model appears more accurate in recognizing arguments of both nominal (88.0\% vs 87.8\% {F\textsubscript{1}}) and verbal predicates (91.2\% vs. 90.5\% {F\textsubscript{1}}). This, when taken together with weaker  labeled {F\textsubscript{1}} of our model for nominal predicates (Table \ref{tab:english-nominal-verbal}), implies that, contrary to our expectations, it is the role labeling performance for nominals which is problematic for our model. 
Examples of this behavior can be seen in  Table \ref{tab:nominal-examples}: all arguments of the predicate {\it pressure} are correctly recognized by our model but  the role for the argument \textit{selling} is not predicted correctly.
In contrast, PathLSTM does not make any mistake with the labeling of the argument \textit{selling} but fails to recognize \textit{from} as an argument.

\section{Related Work}
Earlier approaches to SRL heavily relied on complex sets of lexico-syntactic features \cite{gildea2002automatic}.   
\newcite{DBLP:conf/conll/PradhanHWMJ05} used a support vector machine classifier and relied on two syntactic views (obtained with two different parsers), for feature extraction.
In addition to hand-crafted features, \newcite{DBLP:conf/icml/RothY05} enriched CRFs with an integer linear programming inference procedure in order to encode non-local constraints in SRL; \newcite{DBLP:journals/coling/ToutanovaHM08} employed a global reranker for dealing with structural constraint; while \newcite{DBLP:journals/jair/SurdeanuMCC07} studied several combination strategies of local and global features obtained from several independent SRL models.

In the last years there has been a flurry of work that employed neural network approaches for SRL.
\newcite{fitzgerald-EtAl:2015:EMNLP} used hand-crafted features within an MLP for calculating potentials of a CRF model;
\newcite{roth-lapata:2016:P16-1} extended the features of a non-neural SRL model with LSTM representations of syntactic paths between arguments and predicates;
\newcite{lei-EtAl:2015:NAACL-HLT} relied on low-rank tensor factorization that captured interactions between arguments, predicate, their syntactic path and semantic roles; while \newcite{collobert2011natural} and \newcite{foland2015} used convolutional networks as sentence encoder and a CRF as a role classifier, both approaches employed a rich set of features as input of the convolutional encoder.
Finally, \newcite{DBLP:conf/conll/SwayamdiptaBDS16} jointly modeled syntactic and semantic structures; they extended one of the earliest neural approaches for SRL \cite{HendersonConll08,TitovIjcai09,TitovCoNLL09ST}, with  more sophisticated modeling techniques, for example, using LSTMs instead of vanilla RNNs.

Another related line of work~\cite{naradowsky2012,DBLP:conf/acl/GormleyMDD14}, instead of relying on treebank syntax, integrated grammar induction as a sub-component into their statistical model. In this way, similarly to us, they do not use treebank syntax but rather rely on the ability of their joint model to induce syntax appropriate for SRL. Their focus was primarily on the low resource setting (where syntactic annotation is not available), whereas in standard set-ups their performance was not as strong. It would be interesting to see if explicit modeling of latent syntax is also beneficial when used in conjunction with LSTMs.

\section{Conclusions}

We proposed a neural syntax-agnostic method for dependency-based SRL.
Our model is simple and fast, and surpasses comparable approaches (no system combination, local inference) on the standard in-domain CoNLL-2009 benchmark for English, Chinese, Czech and Spanish. 
Moreover, it outperforms all previous methods (including ensembles) in the arguably more realistic out-of-domain setting in both English and Czech. 
In the future, we will consider integration of syntactic information and joint inference. 

\section*{Acknowledgments}
The project was supported by the European Research Council (ERC StG BroadSem 678254), the Dutch National Science Foundation (NWO VIDI  639.022.518) and an Amazon Web Services (AWS) grant. 
The authors would like to thank Michael Roth for his helpful suggestions.

\bibliography{eacl2017}
\bibliographystyle{acl_natbib}

\end{document}